\documentclass{article}
\usepackage{ijcai19}

\usepackage{times}
\usepackage{soul}
\usepackage{url}
\usepackage[hidelinks]{hyperref}
\usepackage[utf8]{inputenc}
\usepackage[small]{caption}
\usepackage{graphicx}
\usepackage{amsmath}
\usepackage{booktabs}
\usepackage{algorithm}
\usepackage{algorithmic}

\usepackage{helvet}  
\usepackage{courier}  
\usepackage{graphicx}  
\usepackage{booktabs}
\usepackage{array}

\usepackage{color}
\usepackage{mathptmx}
\usepackage{amsmath,amssymb,amsfonts}
\usepackage{bm}
\usepackage{xcolor}
\usepackage{subfigure}

\usepackage{balance}
\usepackage{multirow}

\newcommand{\RNum}[1]{\uppercase\expandafter{\romannumeral #1\relax}}

\newcommand{\hytt}[1]{\texttt{\hyphenchar\font=\defaulthyphenchar #1}}

\title{Super Interaction Neural Networks}

\author{
Yang Yao$^1$
\and
Xu Zhang$^1$\and
Baile Xu$^1$\and
Furao Shen$^1$\and
Jian Zhao$^2$
\affiliations
$^1$Department of Computer Science and Technology, Nanjing University, Nanjing, China\\
$^2$School of Electronic Science and Engineering, Nanjing University, Nanjing, China
\emails
\{zhangxu037,yaoyang,bixu\}@smail.nju.edu.cn,\\
\{frshen,jianzhao\}@nju.edu.cn
}

\begin{document}

\maketitle

\begin{abstract}

Recent studies have demonstrated that the convolutional networks heavily rely on the quality and quantity of generated features.  However, in lightweight networks, there are limited available feature information  because these networks tend to be shallower and thinner due to the efficiency consideration. For farther improving the performance and accuracy of lightweight networks, we develop \textit{Super Interaction Neural Networks (SINet)} model from a novel point of view:  enhancing the information interaction in neural networks. In order to achieve information interaction along the width of the deep network, we propose\textit{ Exchange Shortcut Connection},  which can integrate the information from different  convolution groups without any extra computation cost.  And then, in order to achieve information interaction along the depth of the network, we proposed \textit{Dense Funnel Layer} and \textit{Attention based Hierarchical Joint Decision}, which are able to make full use of middle layer features. Our experiments show that the superior performance of SINet over other state-of-the-art lightweight models in ImageNet dataset. Furthermore, we also  exhibit the effectiveness  and universality of our proposed components by ablation studies. 

\end{abstract}

\section{Introduction}
\label{introduction}
Nowadays, Convolutional Neural Networks (CNNs) are widely applied in various computer vision tasks, including image classification~\cite{Krizhevsky2012alexnet,Huang2016ResNet}, object detection~\cite{Ren2015Faster}, and semantic segmentation~\cite{Chen2018DeepLab}, etc.
In order to achieve higher accuracy, modern CNNs tend to be equipped with hundreds of hidden layers and numerous training parameters~ \cite{Chollet2016Xception,Huang2016ResNet,Huang2016Densely}, resulting in enormous computation loads at billions of FLOPs (\textbf{Fl}oating-Point \textbf{Op}erations).
Therefore, training and deploying deep learning models are still difficult tasks which require plenty of computing resources and expensive hardwares such as GPU or TPU.

Considering the applications in practice, we urgently expect to deploy CNNs on embedded systems such as Internet-of-Things (IOT) devices, mobile phones, etc.
Those embedded systems suffer from computation resource limitation, therefore the general state-of-the-art large CNN models, like ResNet~\cite{he2016resnet} and VGGNet~\cite{simonyan2014very}, are not applicable to them.
For this reason, CNN models with satisfying accuracy and limited computational budgets (tens or hundreds of MFLOPs, i.e. Million Floating-Point Operations) are required for such resource-constrained scenarios.

In recent years, many researches have focused on building lightweight and efficient neural networks, including Xception~\cite{Chollet2016Xception}, MobileNet~\cite{howard2017mobilenets}, MobileNetV2~\cite{Sandler2018MobileNetV2}, ShuffleNet~\cite{zhang2018shufflenet}, etc.
Through comprehensive survey on these studies, we have found that lightweight models tend to be shallower and thinner for efficiency. 
Therefore the features extracted from these networks are limited.
Aiming to achieve satisfying accuracy in such networks, we must make full use of existing feature information.
Thus, different from the focus of lightweight networks design in the past, in this paper, we propose \textit{Super Interaction Network (SINet)}, which pays more attention to how to reduce the network scale by enhancing information interaction.

The contributions of our proposed SINet  are summarized as follows:
\begin{itemize}
\item We propose a novel  feature exchange method in group convolution, namely \textit{Exchange Shortcut Connection}. All convolution operations of SINet are divided into multiple groups for computation reduction, and then we use exchange shortcut connections between groups to enhance the information interaction along the width of network.
\item We propose \textit{Dense Funnel Layer} to achieve the dense connections like DenseNet~\cite{Huang2016Densely} does, but greatly reduce the computational cost to save memory resources. It improves the information interaction along the depth of network.
\item We propose \textit{Attention based Hierarchical Joint Decision}, which comprehensively exploits the usefulness of middle-layer features by means of the attention mechanism.
It improves the information interaction along the depth of network.
\item We experimentally demonstrate the superiority of SINet compared with other state-of-the-art lightweight networks and classic networks on the benchmark datasets,  including ILSVRC-2012 and CIFAR-100.
Specially, we also verify the effectiveness and the universality of our proposed components through elaborate ablation studies.

\end{itemize}

\section{Related Work}
\label{related work}

\subsection{ Efficient Model Design}
There have been many studies focusing on the lightweight network design in recent years.
Two common basic components of these proposed models are \textit{group convolution} and \textit{channel integration}.

\subsubsection{Group convolution}
\label{Group convolution}
Group convolution can save the computational cost by convolving on feature map groups separably, rather than all the feature maps outputted by the preceding layer.  
Therefore, it actually becomes a standard component of lightweight networks.

Group convolution was first introduced in AlexNet \cite{Krizhevsky2012alexnet} for distributing the model over two GPUs to handle the memory explosion issue because of limited hardware resources.
After that, depthwise separable convolution was proposed in Xception~\cite{Chollet2016Xception} and generalized the ideas of group convolution in Inception series.
ResNeXt~\cite{xie2017aggregated} proposed the concept of cardinality which represents the group of channels, and demonstrated the effectiveness of group convolution.
MobileNetV1~\cite{howard2017mobilenets} utilized channel-wise convolution, where each group contains only one channel of feature maps, and achieved state-of-the-art results among lightweight models.
Based on these studies, our SINet also utilizes group convolution through the network.

\subsubsection{Channel Integration}
\label{Channel Integration}
Although group convolution can greatly reduce the computation of the network, it also leads to the information barrier among the groups because each group is independently convoluted.

To deal with this problem, modern CNN library \textit{Cuda-Convnet} supported ``random sparse convolution'' layer, which implemented random channel shuffle after group convolutional layer.
IGCV~\cite{Zhang2017IGCV} adapted the permutation operation in order to mix the  features of different groups.
MobileNet~\cite{howard2017mobilenets} leveraged the point-wise convolution after channel-wise convolution to reintegrate the separated channels.
ShuffleNet~\cite{zhang2018shufflenet} proposed the channel shuffle operation for grouped point-wise convolution.
Different from the previous studies, in this paper, we introduce a novel operation, named Exchange Shortcut Connection, which can be used to share the information among different groups.
Most importantly, it needs no extra computation at all.

\subsection{Network Design and Shortcut Connection}
In order to obtain a more effective network, an intuitive  method is increasing the depth of neural networks~\cite{simonyan2014very}, such as AlextNet~\cite{Krizhevsky2012alexnet} 
and VGGNet~\cite{simonyan2014very}.
However, as the depth of  the network increases, the gradient back propagation becomes more and more difficult, and the gain achieved by simply increasing depth  tends to be saturated.
ResNet~\cite{he2016resnet} proposed  identity shortcut connection to ease this issue of the deep networks.
Based on the shortcut connection  architecture, various models such as WideResNet~\cite{zagoruyko2016wideresnet}, Inception-ResNet~\cite{szegedy2017inception4}, and ResNeXt~\cite{xie2017aggregated} have been developed.
Recently, DenseNet~\cite{Huang2016Densely} proposed a new architecture ---\textit{Dense Connection} which iteratively concatenates the input features with  the output features so that each convolution unit can receive information from all the preceding units.
In this paper, we borrow the idea from the shortcut connection and DenseNet to improve the accuracy and efficiency of our model.

\subsection{Attention Mechanism}
As introduced in ~\cite{Hu2017SENet}, attention mechanism can be interpreted as a method of biasing the allocation of available computational resources towards the most informative components of a signal.
It can be categorized into two classes: \textit{soft attention} that assigns various weights to different components, and \textit{hard attention} that selects a part of components with a certain sampling strategy.

In recent years,  attention mechanism has been applied to a variety of tasks and  proven to be very effective.
These tasks include:
image caption~\cite{lu2017knowing}, machine translation~\cite{luong2015effective}, localization~\cite{Olga2015ImageNet} and so on.
There have also been many attempts~\cite{Hu2017SENet} to incorporate attention mechanism for improving the accuracy in classification tasks.
Residual Attention Network~\cite{wang2017residualattention} proposed an encoder-decoder style attention module to refine the feature mapping.
It makes the network not only works well, but also is robust to noise.
SENet~\cite{Hu2017SENet} introduced Squeeze-and-Excitation module, which achieves the channel attention by global average summarization features.
Inspired by the previous work, in this paper, we propose  Attention based Hierarchical Joint Decision in our lightweight model.
It is used to exploit the features of middle layers for final classification.

\section{Model}
\label{approach}
In this section, we propose three new operations applied in SINet, including  Exchange Shortcut Connection, Dense Funnel Layer and
Attention based Hierarchical Joint Decision.
After that, we illustrate the overall architecture of SINet.

\subsection{Exchange Shortcut Connection}
\label{Exchange Shortcut Connection}
\begin{figure}[h]
\centering
\includegraphics[height=.40\textwidth]{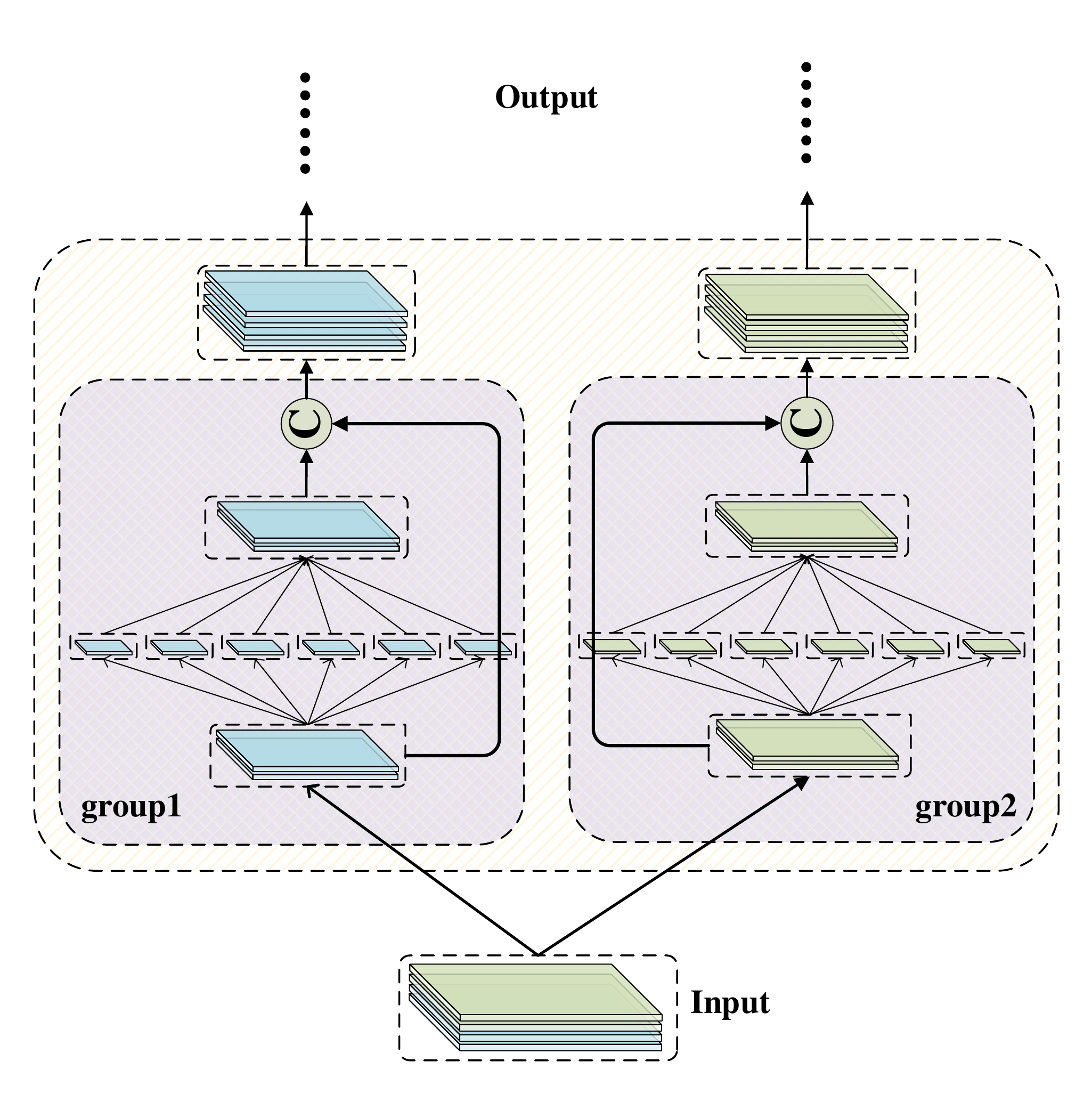}
\caption{Group convolution.  Different colors (blue and green) represent  different groups. Convolution operations are performed independently in both groups.}
\label{fig:group}
\end{figure}

As discussed in Sec.~\ref{Channel Integration}, the group convolution is a commonly-used component in lightweight models.
Our SINet also adopt this operation through the network.
However, although the group convolution can save the computation cost, it is obvious that the information interaction between groups is completely interrupted because the outputs from a certain group are only related to its own inputs, as shown in Figure.~\ref{fig:group}.

Many methods have been proposed to deal with this problem, but they almost require extra structures and computations to mix up separated groups.
For example, in MobileNet, the proposed point-wise convolution for channel reintegration occupies 93.4\% of computation in the network ~\cite{howard2017mobilenets}.

Different from previous work, in SINet, we propose a totally novel and more efficient method, namely \textit{Exchange Shortcut Connection}, which can achieve the information interaction among groups without any extra computation.

\begin{figure}
\centering
\includegraphics[height=.3\textwidth]{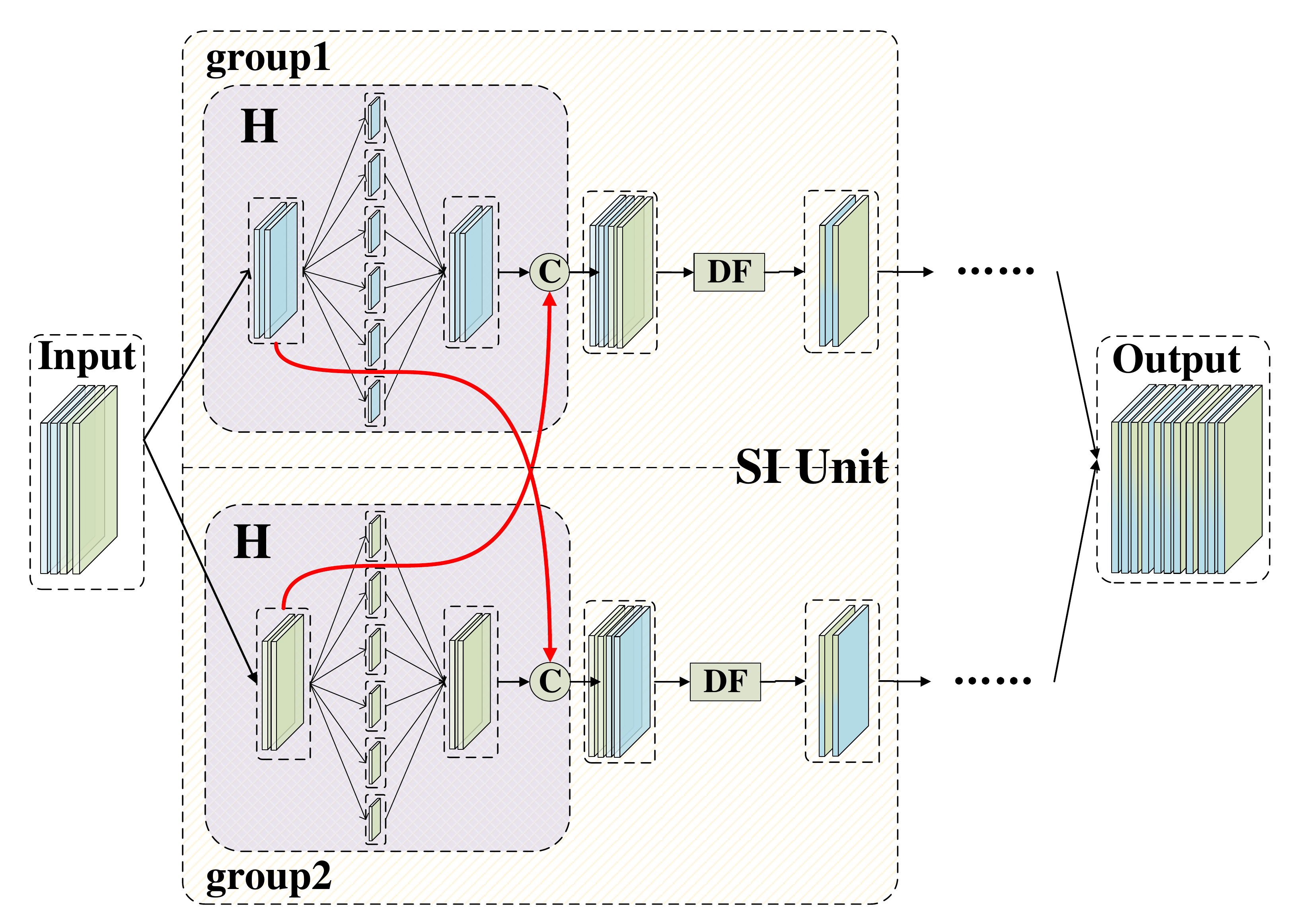}
\caption{The detail of operations in SI Units, where the red bold lines denote the Exchange Shortcut Connections. \textbf{C} denotes concat operation and \textbf{DF} denotes Dense Funnel Layer.}
\label{fig:group and exchange}
\end{figure}

We define $\mathbf{H}(\cdot)$ as a composite convolution operation, and let $\mathbf{H}_l(\cdot), l \in {1, ..., N}$ denote the convolution operation computed by the $l_{th}$  layer.
 $\boldsymbol{X_{0}}$ represents the input image and $\boldsymbol{X_{l}}$ is the output of the $l_{th}$ layer.
Inspired by ResNet~\cite{he2016resnet}, whose standard shortcut connection can be recursively defined as: $\boldsymbol{X_{l}} = \mathbf{H}\left ( \boldsymbol{X_{l-1}} \right )+\boldsymbol{X_{l-1}}$, if we divide convolution operations in our network into two groups ($X_l^1$ and $X_l^2$), Exchange Shortcut Connection can be recursively defined as Eq~\eqref{exchange_resnet}, where $[\cdot]$ means the concatenation operation.
\begin{equation}
\label{exchange_resnet}
\begin{aligned}
&\boldsymbol{X}^1_{l-1},\boldsymbol{X}^2_{l-1}=\mathbf{Split}(\boldsymbol{X}_{l-1})\\
&\boldsymbol{ X}^1_{l} =  \mathbf{H}(\boldsymbol{X}^1_{l-1})+\boldsymbol{X}^2_{l-1}\\
&\boldsymbol{ X}^2_{l} =  \mathbf{H}(\boldsymbol{X}^2_{l-1})+\boldsymbol{X}^1_{l-1}\\
&\boldsymbol{ X}_{l} =[\boldsymbol{ X}^1_{l},\boldsymbol{ X}^2_{l}]
\end{aligned}
\end{equation}
Figure.~\ref{fig:group and exchange} illustrates the group convolutions with Exchange Shortcut Connection.
Compared with Figure.~\ref{fig:group}, Exchange Shortcut Connection allows different convolution groups to access the input data from the others, so that builds up information interaction among groups.
Most importantly, we emphasize that this method needs no extra computational cost and can be easily and flexibly extended to other existing architectures.

Compared with the general CNNs, our structure has less computation complexity under the same settings.
Given a convolution operation with the kernel size $k \times k$, when the input size is $c \times h \times w$ and the  number of the output channels is $m$, general convolution requires $c \times h \times w \times k \times k \times m$ FLOPs, while our proposed operation only requires $c \times h \times w \times k \times k \times m / g$ FLOPs, where $g$ denotes the number of groups.

\subsection{Dense Funnel Layer}

\subsubsection{Dense Funnel Layer}
\label{dense funnel layer}

Traditional shortcut connections~\cite{he2016resnet} only leverage the information of the nearest preceding layer.
To further improve the information interaction along the depth of network, we expect to connect the current layer with all the preceding layers, like dense connection in DenseNet~\cite{Huang2016Densely}.

Although the dense connection has been proven to be very effective in the general CNN models, it would obviously cause a explosive growth of memory and computation cost, which is not acceptable for a lightweight network.
Thus, we introduce a novel structure called \textit{Dense Funnel Layer}.

\begin{figure*}[h!]
\centering
\includegraphics[height=.18\textwidth]{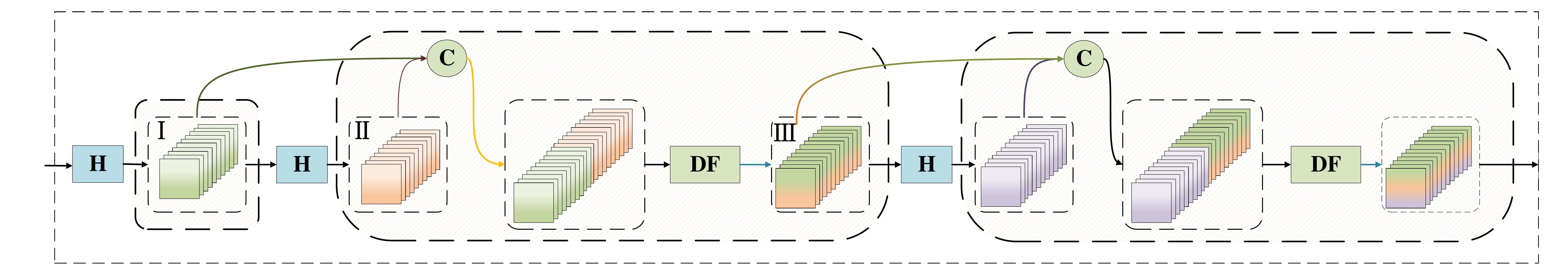}
\caption{An example of Dense Funnel Layer. Where $\mathbf{H}$ denotes Composite function, $\mathbf{DF}$ denotes Dense Funnel Layer and $\mathbf{C}$ denotes concatenation operation.}
\label{fig:narrow}
\end{figure*}

Figure.~\ref{fig:narrow} schematically illustrates the layout of Dense Funnel Layer.
The different colored lines in the figure represent the data flows of different layers.
The green boxes indicate the first set of feature maps at the beginning.
The orange boxes are the feature maps outputted by composition convolution operation $\mathbf{H}$ after applied on green feature maps.

Firstly, we combine input feature maps (green feature maps \RNum{1}) and the current layer output feature maps (orange feature maps \RNum{2}) by concatenating them.
Then we use a $1\times 1$ convolution layer to reduce the channels and mix up the information from both sources, getting the output feature maps \RNum{3}. 
This operation is the key for avoiding the explosive growth of memory caused by the dense connection, because it squeezes out the useless information.
After that, we feed these mixed feature maps to the next layer and conduct similar operations repeatedly.
After several iterations in the network, we can hold a view that the last set of feature maps approximately contains the information of all the preceding feature maps .

\subsection{Attention based Hierarchical Joint Decision}
\label{attention decision}
\begin{figure}[h]
\centering
\includegraphics[width=.5\textwidth]{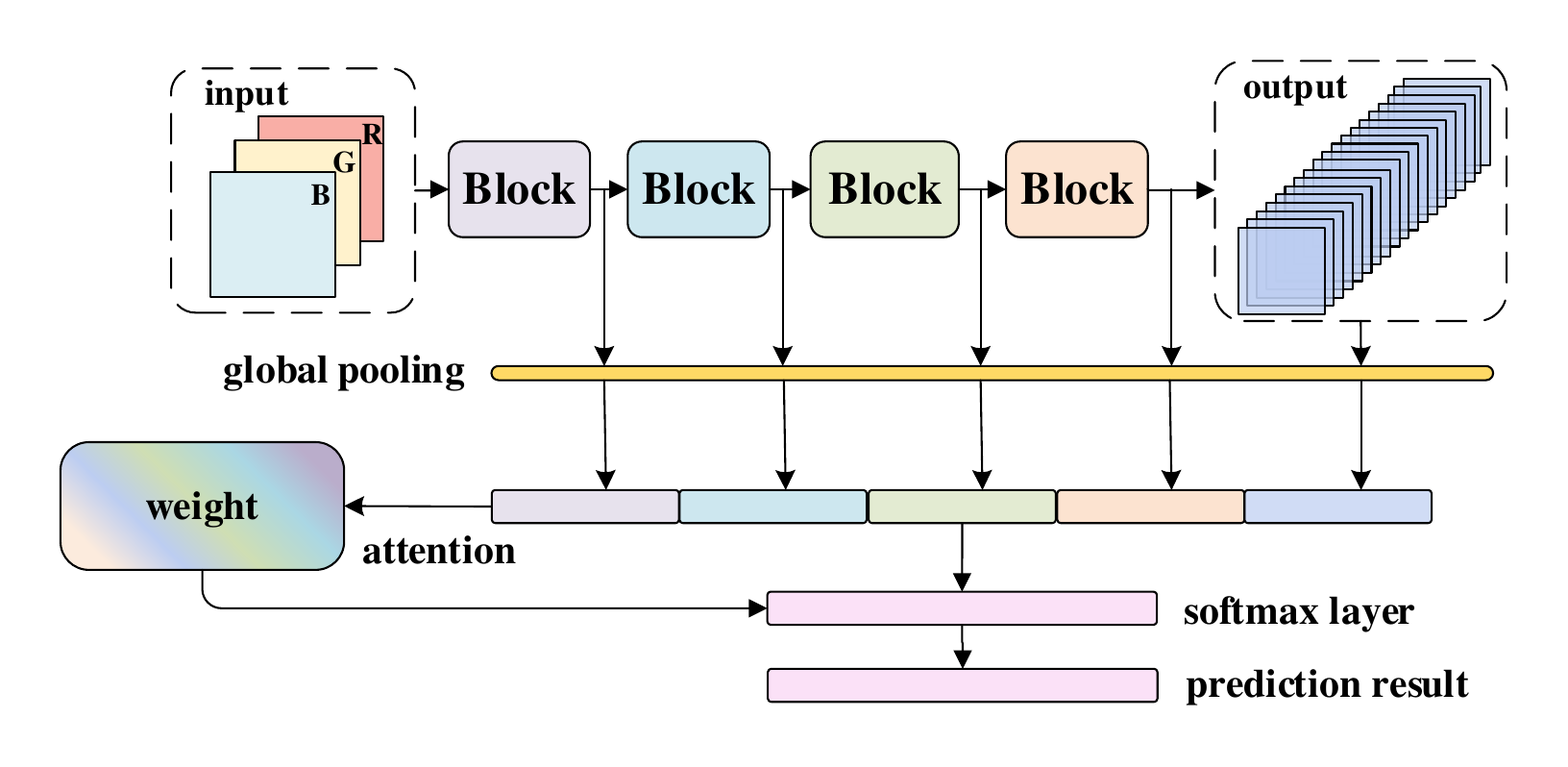}
\caption{Attention based Hierarchical Joint Decision.}
\label{fig:attention}
\end{figure}

High-level features focus more on the semantic meaning and low-level features are more about the edges and textures.
General CNNs conduct classification only based on the last layer of the network which belongs to the high-level features.
Aiming to make full use of existing features in the lightweight network, we utilize the outputs from all the middle layers to conduct classification. 
We call it \textit{Attention based Hierarchical Joint Decision}, as illustrated in Figure.~\ref{fig:attention}.

The most intuitive way of joint decision is to connect the feature maps of different layers directly for classification.
Obviously, this naive idea is only suitable for the same sized feature maps.
However, through the consecutive downsampling operations in the network, the feature maps from different Blocks vary in size.

Recent studies~\cite{Hu2017SENet} have showed that much of the information in convolutional features is captured by the statistics of  different channels and their dependencies.
Thus, we use the average pooling to compress the feature maps $X_{k} \in \mathbb{R}^{h \times w \times c_{k}}$  from Block $k$  into one-dimensional feature vector $Z_{k} \in \mathbb{R}^{1 \times 1 \times c_{k}} $: 
\begin{equation}
Z_{k}=\frac{1}{h \times w} \sum_{i=1}^{h}\sum_{j=1}^{w} x_{i,j,c_k},
\label{equ:avg_pool}
\end{equation} 
and then connect all these feature vectors together.
In this way, we  can avoid the problem of different sizes.

In addition, we notice that the feature maps from different Blocks should be given different attention during joint decision making.
In order to reflect the various importance of different hierarchical features, we endow the feature vectors with adaptive weights by means of attention mechanism.
We add two fully-connected layers on the feature vector $Z_k$, whose output $\alpha_{k}$ is the weight of attention, as shown in Eq~\eqref{equ: ad_weight}.
We expect the network to automatically learn how to assign the attention to feature maps from each level.
\begin{equation}
\alpha_{k}=\sigma((Z_{k}W_{1})W_{2})
\label{equ: ad_weight}
\end{equation}

In Eq.~\eqref{equ: ad_weight},  $\sigma$ refers to the sigmoid activation function , $W_1 \in \mathbb{R}^{ C_{k} \times d}$  and $W_2 \in \mathbb{R}^{d \times 1}$ are the weights, and $d$ is the dimensionality of the hidden layer.

The final classification layer can be computed according to Eq \eqref{eq:attention}, where $W_{cls}$ is the weight of classification layer and $[\cdot]$ denotes the concatenation operation.

\begin{equation}
\label{eq:attention}
pred=\mathbf{softmax}( W_{cls} \cdot  [\alpha_{1} Z_{1} ,..., \alpha_{k} Z_{k} ])
\end{equation}

\subsection{Architecture }

\subsubsection{SI Unit and SI Block}
\label{si unit}

Taking the advantages of  \textit{Exchange Shortcut Connection} and \textit{Dense Funnel Layer} operations, we propose \textit{SI Unit}, in which the amount of computation can be greatly reduced with little accuracy decline.


An SI Unit is composed of three basic  operations, including Exchange Shortcut Connection, Composite Convolution Function $\mathbf{H}$ and Dense Funnel Layer, as illustrated in Figure.~\ref{fig:group and exchange}.

$\mathbf{H}$ consists of a convolution operation~(Conv),  a rectified linear unit (ReLU6) activation function~\cite{Krizhevsky2012alexnet}, and followed by batch normalization (BN)~ \cite{Ioffe2015Batch}.
In the convolution layer, we use a computationally efficient depth-wise convolution~\cite{Chollet2016Xception,howard2017mobilenets} based on the inverse residual bottleneck structure~\cite{Sandler2018MobileNetV2}.

\begin{figure*}[h!]
\centering
\includegraphics[width=\textwidth]{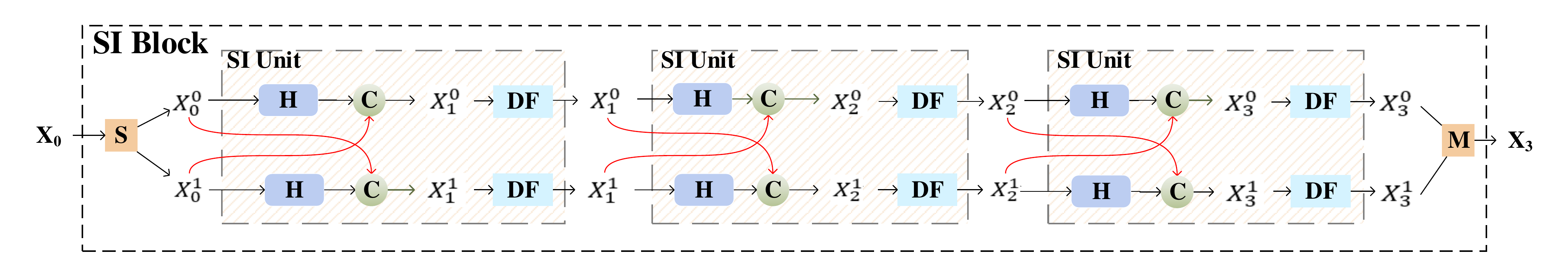}
\caption{An SI Block with three SI Units. \textbf{ S} indicates split operation which divides the input feature maps into $g(g=2)$ groups,\textbf{ M} indicates Merge operation which connects the feature maps of different groups and \textbf{C} indicates concatenation operation in the exchange shortcut.}
\label{fig:exchange units}
\end{figure*}

\textit{SI Block} in our network is formed by a stack of \textit{ SI Units}.
It can achieve full information interaction of the different groups and different units, like shown in Figure.~\ref{fig:exchange units}.
Several SI Blocks are stacked together to construct the entire network.

\subsubsection{Network Architecture Overview}
Built on the SI Units, the SI Blocks and the Attention based Hierarchical Joint Decision, we present the overall architecture of SINet in Table \ref{tab:overall architecture}. 
Similar to MobileNetV2~\cite{Sandler2018MobileNetV2}, the architecture of SINet can be customized by the \textit{width multiplier} parameter $w$ to meet the requirements of accuracy/performance trade-offs.
As $w$ becomes larger, the network suffers from more parameters and computational overhead, but it can achieve higher accuracy.

\begin{table}[h]
\footnotesize
  \centering
  \caption{SINet architecture. Each Block describes a sequence of identical layers, repeated $n$ times. $k$ denotes the kernel size. 
  All layers in the same sequence have the same number of output channels. The first layer of each sequence has a stride $s$ and all others use stride 1. 
  The expansion factor $t$ is the multiplying factor of inverse residual convolution introduced in MobileNetV2, and $w$ is width multiplier.}
    \begin{tabular}{c|c|c|c|c|c|c}
    \toprule
    Layers & output & $k$ & $s$ & $n$ & $t$     & operator \\
    \midrule
    Image & $224^2 \times 3$ &       &       &       &       &  \\
    \midrule
    Block1 & $112^2 \times (24 \cdot w$) & $3 \times 3$    & 2     & 4     & 3     & bottleneck \\
    \midrule
    Block2 & $56^2 \times (40 \cdot w$) & $5 \times 5$    & 2     & 4     & 3     & bottleneck \\
    \midrule
    Block3 & $28^2 \times (80 \cdot w$) & $5 \times 5$    & 2     & 4    & 6     & bottleneck \\
    \midrule
    Block4 & $14^2 \times (96 \cdot w$)  & $3 \times 3$    & 2     & 4    & 6     & bottleneck \\
    \midrule
    Block5 & $7^2 \times (192 \cdot w$) & $5 \times 5$    & 2     & 4    & 6     & bottleneck \\
    \midrule
    \multirow{3}[2]{*}{Decision} & $1^2 \times (432 \cdot w)$ & $7\times 7$     &       & 1     &       & \multicolumn{1}{l}{avg pool} \\
    \cmidrule{2-7}
          & $1^2 \times 1280 $& $1 \times 1 $   &       & 1     &       & fc \\
     \cmidrule{2-7}
          & $1^2 \times$  c&$ 1\times 1$     &       & 1     &       & fc \\
    \bottomrule
    \end{tabular}%
  \label{tab:overall architecture}%
\end{table}%

The architecture of SINet contains a initial fully convolution layer with 21 filters, followed by 5 SI Blocks.
The convolution operation of each  SI Block is performed in two groups, and the first convolution operation of each  SI Block is always applied with stride = 2  for down-sampling.
In the final decision layer, we apply the Attention based Hierarchical Joint Decision operation on  the output feature maps from all SI Blocks.
Then we add a fully-connected softmax layer on it for classification.

\section{Experiment}
\label{experiment}
In this section, we evaluate the effectiveness of SINet through a series of experiments.
Firstly, we exhibit the high accuracy and efficiency of SINet on ImageNet-1000 classification task.
Secondly, we also show the validity and universality of our purposed components by means of Ablation Studies on CIFAR-100 dataset.
The experimental source code can be referred in \hytt{\href{https://github.com/source-code-share/SINet.git}{https://github.com/source-code-share/SINet.git}}.
\subsection{Image Classification on ImageNet}
\subsubsection{Training setup}
\label{dataset:imagenet}
The ILSVRC 2012 classification dataset consists of 1.28 million images for training, and 50,000 images for validation, from 1000 different classes.
We train networks on the training set and report the top-1 and top-5 accuracy on the validation set.
We train our models using Pytorch, and optimization is performed using synchronous Stochastic Gradient Descent (SGD)  algorithm with a momentum of $0.9$ and a batch size of $256$.
Following the MobileNets~\cite{howard2017mobilenets,Sandler2018MobileNetV2} setup, we use a initial learning rate of 0.045, and a learning rate decay rate of 0.98 per epoch.

\subsubsection{Results}
We compare our SINet with several state-of-the-art networks on the ImageNet dataset.
The specific statistics of different models and their accuracies can be seen in Table \ref{tab:imagenet}.

Firstly, compared with some general non-lightweight models,  our SINet can achieve comparable accuracy but only requires about 1/55 FLOPs of VGG-16 and 1/6 FLOPs of ResNet-34.
Secondly, SINet outperforms the state-of-the-art lightweight models (including MobileNets and ShuffleNets) under much lower computation cost.
Finally, we also compare SINet with the models generated by different strategy, such as AutoML and Pruning technology.
NASNet~\cite{Zoph2017nasnet} is the deep networks automatically searched using reinforcement learning models.
CondenseNet~\cite{Gao2017CondenseNet} is an efficient CNNs architecture which encourages feature reuse and prunes filters associated with superfluous features.
SINets can even achieve higher accuracy than them with comparable amounts of FLOPs and parameters, but does not need assistance of extra strategies.
\begin{table}[h]
\footnotesize
  \centering
  \caption{ImageNet comparative experiment results.}
    \begin{tabular}{l|rccc}
    \toprule
    Model   & \multicolumn{1}{c}{Params} & FLOPs & Top1(\%) & Top5(\%)  \\
    \midrule
    VGG-16 & \multicolumn{1}{c}{15.2M} & 1535M     & 71.93    & 90.67     \\
    ResNet-34 & \multicolumn{1}{c}{22.8M} & 3580M     & 74.97      & 92.24      \\
    \midrule
     AlexNet &     \multicolumn{1}{c}{6.1M}     & 720M    & 62.5    & 83.0      \\
    SqueezeNet & \multicolumn{1}{c}{3.2M} & 708M     & 67.5    & 88.2     \\
    MobileNetV1 & \multicolumn{1}{c}{4.2M} & 575M     & 70.6    & 89.5     \\
    MobileNetV2 & \multicolumn{1}{c}{3.6M} & 300M     & 71.8      & 91.0       \\
    MobileNetv2(1.4) & \multicolumn{1}{c}{6.9M} & 585M     & 74.7    & \textbf{92.5}    \\
    ShuffleNet(1.5) & \multicolumn{1}{c}{3.4M} & 292M    & 71.5    & -        \\
    ShuffleNet($\times 2$) & \multicolumn{1}{c}{5.4M} & 524M     & 73.7    & -         \\
    ShuffleNetv2(2$\times$) & \multicolumn{1}{c}{-} & 591M    & 74.9   & -        \\
    \midrule
    \scriptsize{CondenseNet(G=C=4)} & \multicolumn{1}{c}{2.9M} & 274M     & 71.0      & 90.0      \\
   \scriptsize{CondenseNet(G=C=8)} & \multicolumn{1}{c}{4.8M} & 529M     & 73.8    & 91.7    \\
   NASNet-A & \multicolumn{1}{c}{5.3M} & 564M     & 74.0      & 91.3    \\
    \midrule
    SINet(1.0)   &  \multicolumn{1}{c}{ 3.0M}        &   208 M      &  70.1         &  89.4        \\
    SINet(1.2)   &    \multicolumn{1}{c}{ 3.9M}      &  280M       &   71.0      &  90.1       \\
    SINet(1.6)   &    \multicolumn{1}{c}{ 6.0M}      &  468M       & 73.1        &  91.0       \\
    SINet(1.8)   &    \multicolumn{1}{c}{ 7.6M}      &  570M       & \textbf{74.8}         &    \textbf{92.4}      \\
    \bottomrule
    \end{tabular}%
  \label{tab:imagenet}%
\end{table}%

\subsection{Ablation Studies}
\label{ablation}
The core ideas of our proposed SINet lies in SI Units and  Attention based Hierarchical Joint Decision.
In order to verify the universality and effectiveness of both components, we apply them to four different existing models (including two lightweight models and two general CNN models) and compare their accuracy respectively.
For fair comparison, we use the standard implementation of these existing models described in corresponding papers~\cite{howard2017mobilenets,Sandler2018MobileNetV2,Ma2018ShuffleNet,Hu2017SENet}.
We also compare the computation cost of these existing models under different setup.
Following the general testing protocol, we report the amount of parameters and the computational cost   under the standard  input size of  ImageNet dataset  ($224 \times 224$).
\subsubsection{Training setup}
The CIFAR-100 dataset consists of colored natural images with $32 \times 32$ pixels.
The training and testing sets contain 50000 and 10000 images,  respectively.
We train networks on the training set and report the average accuracy of three executions of the experiments on the testing set.  

We train all these models using Pytorch and optimize them using synchronous SGD algorithm with the momentum of $0.9$ and the batch size of $128$.
All the models are trained in 150 epochs from scratch.
The initial learning rate is set to $0.01$ and decreased by a factor of 10 every 80 epochs.

\subsubsection{SI Unit}
In order to verify the effectiveness of the SI Unit, we compare the original networks \textbf{A} without group convolution ($G=1$), the networks \textbf{B} that only use group operation ($G=2$) but no shortcut exchange connection (EX=No), and the networks \textbf{C} that utilize the SI Unit, including both group convolution  and shortcut exchange connection ($G=2$, EX=Yes).
For fairness, we use exactly the same settings to train these models.
Results are shown in Table \ref{tab:exchange unit}.

\begin{table}[h]

\footnotesize
  \centering
  \caption{Ablation Study on SI Unit under different networks architecture. $G$ denotes the number of groups, and $EX$ denotes whether to use Exchange Shortcut Connection.}
    \begin{tabular}{c|ccccc}
    \toprule
    Model   & $G$       & $EX$      & Params(M) & FLOPs(M) & Acc.(\%) \\
    \midrule
    \multirow{3}[1]{*}{MobileNet} & 1       & -       & 4.23    & 575.00     & 74.33    \\
        & 2        & No       & 2.66    & 291.55  & 73.95    \\
            & 2         & Yes       & 2.66    & 291.55  & \textbf{75.51}    \\
    \midrule
    \multirow{3}[1]{*}{MobileNetv2} & 1       & -       & 3.56    & 322.47  & 71.59    \\
        & 2        & No       & 2.89    & 224.38  & 70.74    \\
            & 2    &Yes       & 2.89    & 224.38  & \textbf{71.71}    \\
    \midrule
    \multirow{3}[1]{*}{ResNet34} & 1       & -       & 21.79   & 3579.76 & \textbf{75.90}    \\
        & 2       & No      & 12.99   & 2109.00    & 73.22   \\
            & 2       & Yes      & 12.99   & 2109.00    & \textbf{74.35}   \\
    \midrule
    \multirow{3}[1]{*}{SENET} & 1       & -       & 11.79   & 1730.16 & \textbf{75.77}   \\
        & 2        & No      & 7.37    & 978.68  & 73.80     \\
           & 2       & Yes      & 7.37    & 978.68  & \textbf{75.23}    \\
    \bottomrule
    \end{tabular}%
  \label{tab:exchange unit}%
\end{table}%

From the results, we can see that the networks \textbf{B} which only use group operation can greatly reduce the amount of parameters and computation, but at the same time, its accuracy is lower than that of original network \textbf{A} in all four network architecture.
The networks \textbf{C} using the SI Unit can achieve the same amount of parameters and computational complexity as the networks \textbf{B} while maintaining a satisfying accuracy compared with the original networks \textbf{A}.
Especially in lightweight  networks (MobileNetV1 and MobileNetV2), the networks \textbf{C} using the SI Unit even achieve better results than that of the original networks \textbf{A}.
It shows that SI Unit compensates for information blocking due to the group convolution.
In addition, it also indicates that the information interaction in shallow networks is more important than that in deep networks.

\subsubsection{Attention based Hierarchical Joint Decision }
In order to evaluate the validity of the Attention based Hierarchical Joint Decision, we compare the original networks \textbf{A} which use general classification layer (Attention=No) and the networks \textbf{B} that utilizes the Attention based Hierarchical Joint Decision  (Attention=Yes).
We use the same settings as the former experiment to train these models.
Results are shown in Table \ref{tab:attention cifar100}.
\begin{table}[h]

\footnotesize
  \centering
  \caption{Ablation Study on Attention based Hierarchical Joint Decision under different networks architecture.}
    \begin{tabular}{c|cccc}
    \toprule
    Model   & Attention      & Params(M) & FLOPs(M) & Acc.(\%) \\
    \midrule
    \multirow{2}[1]{*}{MobileNet} & No       & 4.23    & 567.72  & 74.33 \\
           & Yes       & 5.24    & 567.72  & \textbf{74.66} \\
    \midrule
    \multirow{2}[1]{*}{MobileNetv2} & No       & 3.56    & 322.47  & 71.59 \\
            & Yes       & 2.43    & 322.47  & \textbf{73.24} \\
    \midrule
    \multirow{2}[1]{*}{ResNet34} &No       & 21.79   & 3579.76 & 74.48 \\
                & Yes       & 22.32   & 3579.76 & \textbf{76.06} \\
    \midrule
    \multirow{2}[1]{*}{SENET} & No       & 11.79   & 1730.16 & 75.45 \\
            & Yes       & 12.30    & 1730.16 & \textbf{75.50} \\
    \bottomrule
    \end{tabular}%
  \label{tab:attention cifar100}%
\end{table}%

We can see that networks \textbf{A} using Attention based Hierarchical Joint Decision consistently achieve better performance than the original networks \textbf{B}, at the cost of a small amount of additional computations. 
It shows that the Attention based Hierarchical Joint Decision  is effective and also demonstrates that effective information interaction  along the depth of network is necessary.
\section{Conclusion}
\label{conclusion}
In this paper, we propose a novel lightweight convolutional neural network named SINet. 
In SINet,  for  information  interaction  along  the width of network, we propose Exchange Shortcut Connection,  which  can  integrate  the  information from different convolution groups without any extra computation cost. 
In addition, in order to achieve information interaction along the depth of network, we proposed Dense Funnel Layer and Attention based Hierarchical Joint Decision.
Through experiments on the ImageNet dataset, we demonstrate that SINet outperforms the state-of-the-art lightweight models and widely-used general networks like VGG-16.
Ablation studies on the CIFAR-100 datasets confirms the universality of our newly designed components in SINet.

\bibliographystyle{named}
\bibliography{ijcai19}

\end{document}